\documentclass[
]{ceurart}

\usepackage{hyperref}
\usepackage{tabularx}

\begin{document}

\copyrightyear{2022}
\copyrightclause{Copyright for this paper by its authors.
  Use permitted under Creative Commons License Attribution 4.0
  International (CC BY 4.0).}

\conference{DE-FACTIFY :Workshop on Multi-Modal Fact Checking and Hate Speech Detection, co-located with AAAI 2022. 2022
Vancouver, Canada}

\title{GPTs at Factify 2022: Prompt Aided Fact-Verification}
\author[1,2]{Saksham Aggarwal}[%
email=sakshamaggarwal20@gmail.com,
]
\address[1]{Indian Institute of Technology (Indian School of Mines), Dhanbad, India}
\address[2]{The authors contributed equally.}

\author[1,2]{Pawan Kumar Sahu}[%
email=pawankumar.s.2001@gmail.com,
]
\author[1,2]{Taneesh Gupta}[%
email=tanishgupta34@gmail.com,
]
\author[1,2]{Gyanendra Das}[%
email=gyanendralucky9337@gmail.com,
]

\begin{abstract}
One of the most pressing societal issues is the fight against false news. The false claims, as difficult as they are to expose, create a lot of damage. To tackle the problem, fact verification becomes crucial and thus has been a topic of interest among diverse research communities. Using only the textual form of data we propose our solution to the problem and achieve competitive results to other approaches. We present our solution based on two approaches - PLM (pre-trained language model) based method and Prompt based method. PLM based approach uses the traditional supervised learning, where the model is trained to take 'x' as input and output prediction 'y' as P(y$|$x). Whereas, Prompt-based learning reflects the idea to design input to fit the model such that the original objective may be re-framed as a problem of (masked) language modelling. We may further stimulate the rich knowledge provided by PLMs to better serve downstream tasks by employing extra prompts to fine-tune PLMs. Our experiments showed that the proposed method performs better than just fine tuning PLMs. We achieved an F1 score of 0.6946 on the FACTIFY dataset and 7$^{th}$ position on the competition leader-board.

\end{abstract}


\maketitle

\section{Introduction}
When it comes to news consumption, social media has two faces. On one hand, individuals consume news via social media because of its low cost, quick access, and rapid transmission of information. On the other side, it facilitates the widespread circulation of fake news, i.e., low-quality news containing false/misleading information. As per a study, Facebook engagements with fake news sites average roughly 70 million per month \cite{zeng2021automated}. It impacts government, media, individuals, health, law and order, as well as the economy. This spread has been blamed for incidents ranging from ethnic violence, inter-racial violence, and religious conflicts to mass riots. Thus, combating fake news is one of the burning societal crisis.

Our work in the competition tries to deal with the above mentioned fact-checking problem using the Factify dataset. FACTIFY is a multi-modal fact verification dataset consisting of five classes- “Support-Multimodal", "Support-Text", "Insufficient-Multimodal", "Insufficient-Text" and "Refute" - categorized based on the cross-relationship of visual and textual data (see section 3). Though the support of visual data was provided, our solution uses only textual information and proposes an approach to achieve competitive results with the solutions that leverages visual information.

In a standard supervised learning practice for NLP, we take an input 'x', generally text, and output prediction 'y' based on a model P(y|x) . 'y' might be a label, a text string, or any other type of output. we utilise a dataset comprising of pairs of inputs and outputs, to learn the parameters of this model, and then train to predict the conditional probability. This is generally done by following a pretrain-finetune strategy with additional task specific data using pretrained language models (PLMs).

Language models that directly estimate the likelihood of text are the basis for prompt-based learning. To use these models to perform prediction tasks, the original input 'x' is converted into a textual string prompt 'xprompt' having some unfilled slots using a template, and then the language model is used to probabilistically fill the unfilled information to produce a final string, from which the final output 'y' can be derived.

We model our solution based on both the approaches, and have shown a way to use the prompt-based learning technique to aid in the classification. First approach - traditional pretrain-finetune based method (also referenced as PLM based method further) - focuses on fine tuning different pretrained models with some pre-processing, later aggregated using ensembling techniques. Our second approach - Prompt based - divides the 5-class classification problem into two parts, first, a binary classification task to efficiently segregate one of the classes from other 4, and second, a 4-class classification task which is handled similar to the first approach. We observe that merging the prompt based technique with traditional approach, boosts the score by aiding in efficient segregation of one of the classes. Task report for Factify 2022 can be found here \cite{task_Report}.

\section{Related Work}
Pretrained language models (PLMs) such as RoBERTa \cite{liu2019roberta}, GPT \cite{Radford2018ImprovingLU}, T5 \cite{raffel2020exploring} and BERT \cite{devlin2019bert} have proven themselves as powerful tools for text generation and language understanding. These models are capable of capturing a plethora of semantic \cite{yenicelik-etal-2020-bert}, linguistic \cite{jawahar-etal-2019-bert} and syntactic \cite{hewitt-manning-2019-structural} knowledge leveraging large-scale corpora that's now available to us. Finetuning these PLMs by introducing additional task specific data, rich knowledge in the language models can be propagated to multiple downstream tasks. Demonstration of outstanding performance on nearly all key language tasks, by just finetuning pretrained language models, has upraised a consensus in the community to finetune PLM instead of learning models from scratch \cite{2020}. 

Despite the effectiveness of fine-tuning pre-trained language models, several recent research have discovered that one of the most significant challenges is the substantial gap in the 'pretraining' and 'fine-tuning' objectives. This restricts taking full advantage of knowledge in PLMs. Although 'pre-training' is typically formalized as a cloze-style task (e.g. MLM), downstream tasks in 'fine-tuning' exhibit different objective forms such as sequence labelling, classification and generation. This discrepancy obstructs the transfer and adaption of PLM knowledge to downstream tasks.

\cite{brown2020language} proposes prompt tuning to bridge this gap between pretraining strategy and subsequent finetuning and downstream tasks. To understand the basic idea, we can consider the example of dance performance classification task , see figure 1 for reference, a typical prompt would consists of a template (e.g. “$<$Statement$>$. The guests appreciated the dance performance. It was [MASK].”) and a set of label words (e.g. “outstanding” and “worse”). The set of label words serves as the constituent set for predicting [MASK]. This way, the original input is modified with the help of prompt template to predict [mask] which is then mapped to the corresponding labels, thereby converting a classification task into a cloze-style task.
Simply put, we can make the model predict “outstanding” or “worse” using PLMs at the position which is being masked, which is then used to derive the sentiment (i.e positive or negative). Prompt tuning methods have also achieved promising results on some other few-class classification tasks as well such as natural language inference \cite{schick2021exploiting}.

Taking inspiration from both these approaches, i.e. Pre-train$\rightarrow$ Fine-tune and Pre-train $\rightarrow$ Prompt$\rightarrow$ Predict, we present our solution for Multi-Modal Fact Verification dataset based on both the approaches.

\begin{figure}
  \centering
  \includegraphics[width=\linewidth]{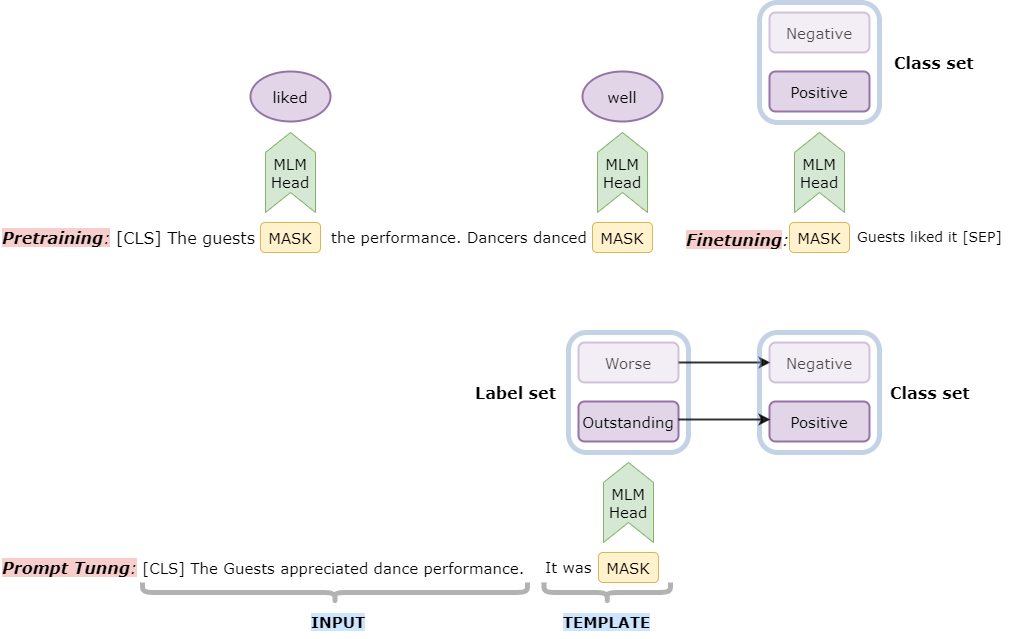}
  \caption{An example of pre-training, fine-tuning, and prompt tuning.}
\end{figure}

\section{Dataset}
Factify is a datatset on multi-modal fact verification \cite{dataset_factify}. It contains images of the claim, textual claim, reference textual document and image. The images are accompanied by their respective OCR texts. The train data includes 35,000 instances maintaining a proper balance among all the five classes - 'support-text', 'insufficient-multimodal', 'support-multimodal', 'insufficient-text' and 'refute'. The validation dataset contains 7,500 instances with equal data distribution across the classes. Short description of the classes is provided in table \ref{tab:classes description}.

\begin{table}[!ht]
\begin{center}
\centering
\caption{The table gives the description of the classes of the dataset}{\centering}
\label{tab:classes description}
\begin{tabularx}{\textwidth} { 
  | >{\raggedright\arraybackslash}X 
  | >{\centering\arraybackslash}X 
  | >{\centering\arraybackslash}X | c | }
 \hline
 \textbf{Support Multimodal} & Text is entailed & Image is entailed \\
 \hline
  \textbf{Support Text} & Text is entailed & Image is not entailed \\
  \hline
 \textbf{Insufficient Multimodal} & Text is not entailed & Image is entailed \\
  \hline
 \textbf{Insufficient Text} & Text is not entailed & Image is not entailed \\
  \hline
 \textbf{Refute} & Fake Claim & Fake Image \\
\hline
\end{tabularx}

\end{center}
\end{table}

\section{Method}

In our solution modelling we have used pretrained RoBERTa, DeBERTa, XLM-RoBERTa and ALBERT models.

\textbf{RoBERTa} \cite{liu2019roberta} (Robustly Optimized BERT Pre-training Approach) improves on BERT \cite{devlin2019bert} by modifying and optimizing its architecture and training procedure. It makes some key changes to BERT including the removal of Next Sentence Prediction (NSP) objective. RoBERTa also dynamically changes the masking pattern by duplicating the training data and masking it 10 times, each time with a different strategy. It uses larger mini batches for training which improves perplexity on masked language modelling objective and also makes it easier to parallelize via distributed data parallel training. 

\textbf{DeBERTa} \cite{he2021deberta} (Decoding-enhanced BERT with disentangled attention) is a Transformer-based neural language model pretrained which improves on BERT and RoBERTa by using two novel techniques. It makes use of disentangled self-attention which, unlike BERT, uses two vectors to encode word(content) embedding and positional embedding instead of using their summation as a single vector. Secondly, it proposes Enhanced Mask Decoder(EMD), which takes into account both relative as well as absolute position of the words.

\textbf{XLM-RoBERTa} \cite{conneau2020unsupervised} is a pretrained multilingual model version of RoBERTa which outperform multilingual BERT. It is pretrained using Masked language modeling objective on 2.5TB of filtered CommonCrawl data including 100 languages due to which it is capable of giving result in 100 diffrerent languages.

\textbf{ALBERT} \cite{lan2020albert} is a Transformer architecture based on BERT which reduces its model size (18x fewer parameters) without deteriorating the performance. It uses 2 parameter reduction techniques including Factorized embedding parameterization and Cross Layer parameter sharing. ALBERT showcases excellent trade-off between huge size reduction and slight performance drop.

\begin{table*}[t]
\begin{center}
    \caption{The table shows the validation scores for the various models trained and along with the score obtained by performing stacking over these models.}
    \label{tab:Validation-result-description}
\begin{tabularx}{1.0\textwidth}{ 
   >{\centering\arraybackslash}X 
   >{\centering\arraybackslash}X  }
 \hline
  \hline

 \textbf{Model-Name} & \textbf{Validation-Score} \\
 \hline 
 DeBERTa & 0.7305\\
 RoBERTa & 0.7082\\
 XLM-RoBERTa & 0.6985\\
 ALBERT & 0.6871\\
 Stacking Ensemble & 0.7360\\
 \hline
  \hline

\end{tabularx}
\end{center}

\end{table*}

\subsection{Data Preprocessing}
The train data consists of claim text, claim OCR text, document text and document OCR text. For training, we concatenated claim text with OCR text and clipped the max length to 256. We tried training on document text as well, but the results were poor. We used stratified 5-fold cross validation strategy for training all of our models.

\subsection{PLM Based Method}
In this method we have used pretrained RoBERTa, DeBERTa, XLM-RoBERTa and ALBERT models and finetuned it on the given dataset. This approach uses the traditional supervised learning, which trains a model to take in an input x and predict an output y as P(y$|$x), here, x denotes the textual data, y denotes the class set and P(y$|$x) denotes the probability. Later the predictions from all the four models is ensembled to boost the final score. The validation scores are mentioned in table \ref{tab:Validation-result-description}. The training strategy along with the ensembling details is described in the Experiments section.

\subsection{Prompt Based Method}
This approach makes use of the 'prompt-based learning' methodology to aid in the classification. A prompt consists of a template T(·) and a set of label words V. For each instance x, the template is
first used to map x to the prompt: xprompt = T(x). Given v $\in$ V, we
produce the probability that the token v can fill the masked position. Then the predicted token v is mapped to its defined class. This step known as answer mapping, maps the distribution over label set(V) to a distribution over class set(Y).

Using this technique we filter out the 'refute' class more efficiently, leveraging the rich knowledge distributed in the pretrained models. We transformed the multi-class classification task to binary classification where the 'refute' class would represent negative class and all other classes ('support-text', 'insufficient-multimodal', 'support-multimodal' and 'insufficient-text') combined would represent the positive class. 

The template we used in our case was " $<$INPUT$>$. The statement is $<$MASK$>$ ". And, the Label set (V) was comprised of words like 'false', 'irrelevant', 'incorrect' etc mapped to the negative class of Y and words like 'true', 'relevant', 'correct' etc mapped to the positive class of Y. The model used was pretrained RoBERTa. This prompt setting was motivated by the nature of 'refute' class and the fact that prompt-based learning is efficient for binary classification. Once the refute class' instances are segregated, now the task at hand becomes a multi-class classification with 4 classes instead of 5. Further, the task is completed using traditional supervised learning - PLM based approach.

\section{Experiments} 

\subsection{Pretraining}
To pretrain our baseline, we employed the Masked Language Model (MLM) method, which substitutes a special masked token for randomly picked tokens in the input sequence. Using BCE loss, the MLM attempts to predict the masked tokens. 15\% of the input tokens were chosen uniformly for replacement, with masked tokens replacing 80\% of the tokens, randomly selected vocabulary token replacing 10\%, and rest 10\% were kept unmodified. AdamW is used to optimise it, using an L2 weight decay of 0.01.The learning rate is warmed up to a peak value of 1e-4 over the first 500 iterations, then linearly decreased.

Models are pre-trained for 3000 iteration with a mini-batch size of 64 and a maximum length of 256.

\subsection{Finetuning}
We used the pretrained models from above and finetuned them on the given dataset using AdamW as the optimizer with weight decay of 0.01. The learning rate is warmed up over 100 iterations to peak value of 5e-6 and decayed using cosine annealing. Models are finetuned for 2000 iterations with mini-batch size 32 of maximum length = 256 tokens.

\subsection{Ensembling}
We used the following ensembling techniques which gave a significant boost to our scores.
\subsubsection{Snapshot}
Training multiple deep models for bagging requires heavy computation. By using snapshot ensembling\cite{huang2017snapshot}, we can obtain different models by training our model only once, converging to several local minima along its optimization path. We can make an ensemble of these multiple models to make more general purpose predictions and thereby boosting the system's performance. To obtain repeated convergence we used cyclic learning rate schedules e.g. OneCyclicLR scheduler.

\subsubsection{Stacking}
Out of fold predictions were predicted for all the models i.e. when performing 5-fold cross-validation training strategy, we predicted the validation scores for all the folds which were then concatenated. These out of fold predictions can be used for ensembling. A 3 layer neural network was trained on these out of fold predictions to predict the target labels. This neural network was then used as an head on the test predictions generated by the transformer based models. This method, known as stacking, gave a significant boost in our validation score (0.7360).

\section{Result}
We ranked 7$^{th}$ in the final leaderboard. Our results are mentioned in table \ref{tab:Results description}. The solution consists of two different approaches. Method 1 refers to the stacking of the finetuned models including DeBERTa, RoBERTa, XLM-RoBERTa and ALBERT while Method 2 refers to the prompt-based approach explained in the method section. Method 2 gave us better results on the final testing dataset owing its boost to the prompt based learning.

\begin{table*}[!ht]
\begin{center}
\caption{Table shows the results based on our two methods. Method 1 refers to stacking of the finetuned models. Method 2 refers to the prompt based approach. }
\label{tab:Results description}
\begin{tabularx}{1.0\textwidth}{ 
  | >{\centering\arraybackslash}X 
  | >{\centering\arraybackslash}X 
  | >{\centering\arraybackslash}X
  | >{\centering\arraybackslash}X
  | >{\centering\arraybackslash}X
  | >{\centering\arraybackslash}X
  | >{\centering\arraybackslash}X | }
 \hline
 \textbf{Method} & \textbf{Support-Text} & \textbf{Support-Multimodal} & \textbf{Insufficient-Text} & \textbf{Insufficient-Multimodal} & \textbf{Refute} & \textbf{Final}\\
 \hline 
 Method1 & 0.726 & 0.7912 & 0.7437 & 0.7755 & 0.996 & 0.6902\\
 \hline
 Method2 & 0.715 & 0.7903 & 0.7536 & 0.7927 & 1.0 & 0.6946\\
 \hline
\end{tabularx}
\end{center}
    
\end{table*}

\section{ Conclusion}
The task was to classify the given claim into one of the 4 labels-  'support-text', 'insufficient-multimodal', 'support-multimodal', 'insufficient-text' and 'refute'. We followed 2 approaches for this task: 1) finetuning transformer based models on our dataset and 2) using prompt based learning to classify if the claim is a refute and then using transformer based models for the downstream task. We carefully evaluated different methods that we used and we found out that the prompt based method was giving us a significant boost. Pretraining using MLM on the given dataset was also giving a small boost compared to finetuning of pretrained model. We also found that pretraining using MLM for more iteration with bigger batch size and dynamically changing the masking pattern applied to the training data was improving the score. Snapshot ensemble of various checkpoints also gave a boost to our single model score. Performing stacking on our prediction further improved the results and was outperforming the mean ensemble of our predictions.

\bibliography{egbib}

\end{document}